\documentclass{article}

\usepackage[nonatbib,preprint]{neurips_data_2024}

\usepackage[utf8]{inputenc} 
\usepackage[T1]{fontenc}    
\usepackage{hyperref}       
\usepackage{url}            
\usepackage{booktabs}       
\usepackage{amsfonts}       
\usepackage{nicefrac}       
\usepackage{microtype}      
\usepackage{xcolor}         
\usepackage{graphicx}       
\usepackage{amsmath}
\usepackage{wrapfig}
\usepackage[style=numeric,sorting=none]{biblatex}       
\usepackage{appendix}
\usepackage{booktabs}
\usepackage{multirow}
\usepackage{epsfig}
\usepackage{colortbl}
\usepackage{xcolor}
\usepackage{makecell}

\newcommand{\chenyi}[1]{\textcolor{black}{#1}}

\title{
EgoPlan-Bench: Benchmarking Multimodal Large Language Models for Human-Level Planning
}

\author{\textbf{Yi Chen}$^{1,2}$, \textbf{Yuying Ge}$^{1*}$, \textbf{Yixiao Ge}$^{1,3}$, \textbf{Mingyu Ding}$^4$, \textbf{Bohao Li}$^{1}$, \textbf{Rui Wang}$^{1}$,\\
\textbf{Ruifeng Xu}$^{5}$, \textbf{Ying Shan}$^{1,3}$, \textbf{Xihui Liu}$^2$\thanks{Corresponding authors.}\\
    $^1$Tencent AI Lab,
    $^2$The University of Hong Kong,
    $^3$ARC Lab, Tencent PCG, \\
    $^4$University of California, Berkeley, 
    $^5$Peng Cheng Laboratory\\
  \href{https://chenyi99.github.io/ego_plan/}{\small https://chenyi99.github.io/ego\_plan/}
 }

\begin{document}

\maketitle

\begin{wrapfigure}{r}{0.47\textwidth}
\vspace{-1mm}
    \centering
    \includegraphics[scale=0.48]{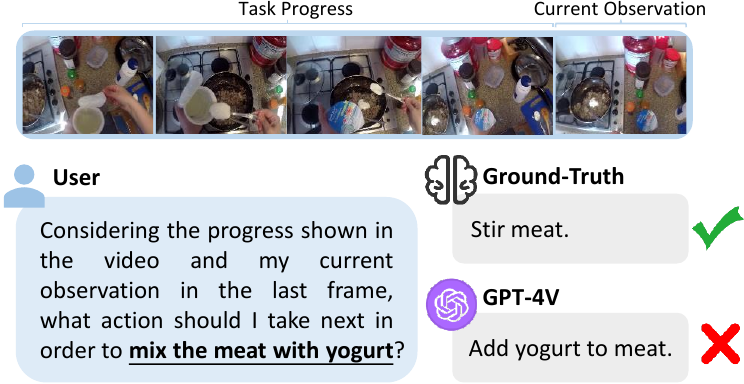}
    \caption{EgoPlan-Bench poses significant challenges for achieving human-level planning with MLLMs (even GPT-4V).
    }
    \label{fig:gpt4v_failure_Case}
\end{wrapfigure}

\begin{abstract}
The pursuit of artificial general intelligence (AGI) has been accelerated by Multimodal Large Language Models (MLLMs), which exhibit superior reasoning, generalization capabilities, and proficiency in processing multimodal inputs. A crucial milestone in the evolution of AGI is the attainment of human-level planning, a fundamental ability for making informed decisions in complex environments, and solving a wide range of real-world problems. Despite the impressive advancements in MLLMs, a question remains: \textbf{How far are current MLLMs from achieving human-level planning?}
To shed light on this question, we introduce EgoPlan-Bench, a comprehensive benchmark to evaluate the planning abilities of MLLMs in real-world scenarios from an egocentric perspective, mirroring human perception. EgoPlan-Bench emphasizes the evaluation of planning capabilities of MLLMs, featuring realistic tasks, diverse action plans, and intricate visual observations. Our rigorous evaluation of a wide range of MLLMs reveals that EgoPlan-Bench poses significant challenges, highlighting a substantial scope for improvement in MLLMs to achieve human-level task planning. To facilitate this advancement, we further present EgoPlan-IT, a specialized instruction-tuning dataset that effectively enhances model performance on EgoPlan-Bench. We have made all codes, data, and a maintained benchmark leaderboard available at \href{https://github.com/ChenYi99/EgoPlan}{https://github.com/ChenYi99/EgoPlan} to advance future research.
\end{abstract}

\section{Introduction}
\label{sec:intro}
\begin{figure*}[!t]
    \centering
    \includegraphics[width=1\textwidth]{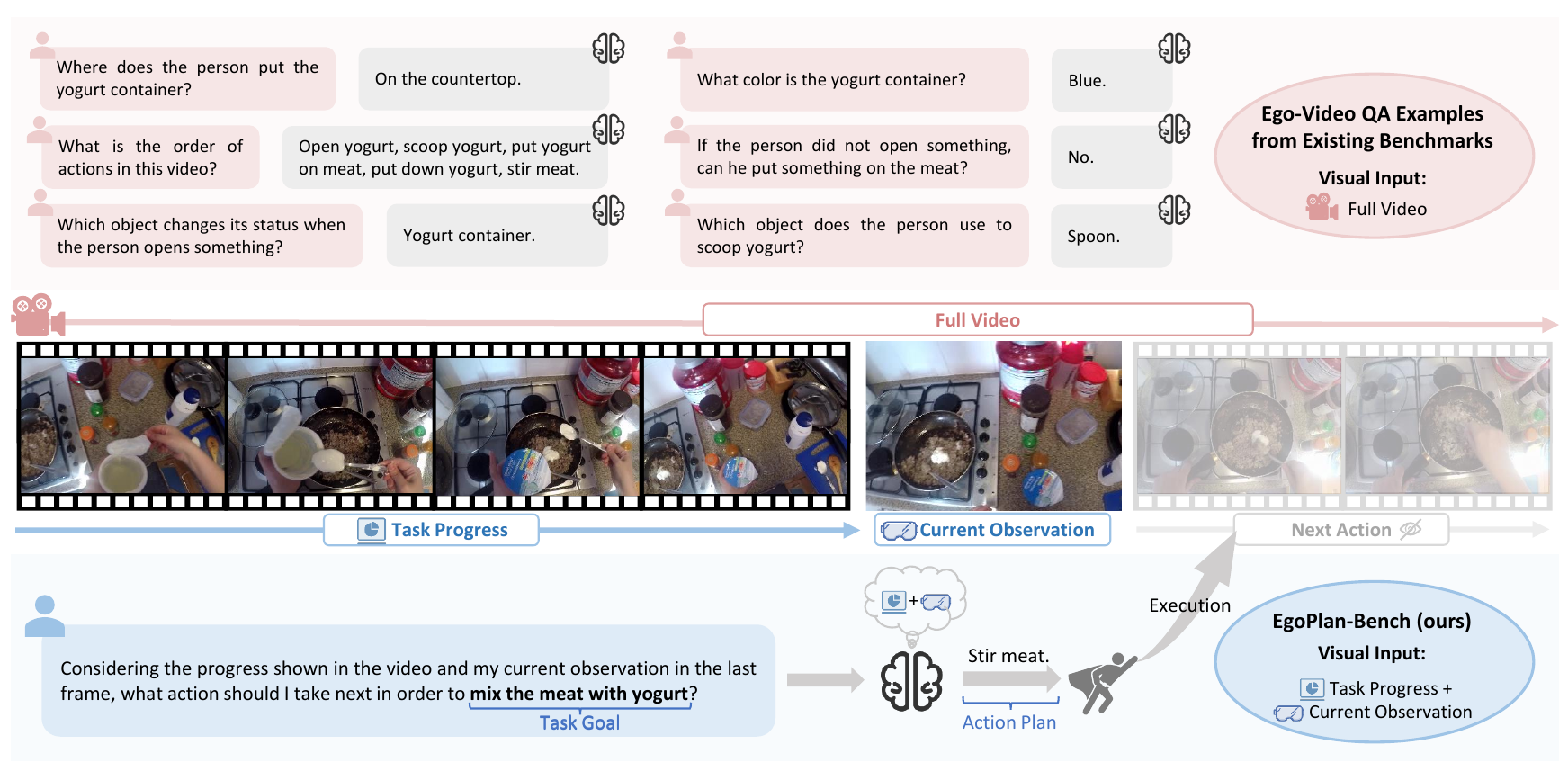}
    \caption{
    Our EgoPlan-Bench evaluates \textbf{Planning}, where a model predicts the next feasible action plan by taking a video showing task progress, current visual observation, and open-form task goal as inputs like humans. In contrast, the egocentric-video-based QA examples from existing benchmarks mainly evaluate \textbf{Comprehension}, where a model answers questions based on the spatial and temporal understanding of the entire video. 
    }
    \label{fig:benchmark_overview}
\end{figure*}
        
Multimodal Large Language Models (MLLMs) combine the exceptional reasoning and generalization capabilities of Large Language Models~\cite{brown2020gpt3, ouyang2022training_instructGPT, touvron2023llama, chiang2023vicuna} with the capacity to understand multimodal inputs. This integration has opened new possibilities for achieving the ultimate goal of artificial general intelligence (AGI), which aims to match or surpass human performance in most tasks~\cite{y2023artificial,bubeck2023sparks,morris2023levels}. A crucial milestone in the evolution of AGI is the achievement of human-level planning capabilities, as task planning constitutes an essential aspect of everyday human activities, enabling informed decision-making in complex environments and the resolution of numerous real-world problems.

Task planning involves predicting the subsequent action plan based on the task goal, real-time progress, and current observations, as illustrated in Figure \ref{fig:benchmark_overview}. This capability is indispensable for a model to be effectively applied in the real world for addressing complex problems, in the same manner as humans. It should be regarded as one of the fundamental criteria for determining whether MLLMs have genuinely attained human-level intelligence. 
However, achieving human-level planning for MLLMs is far from trivial.  
For example, as shown in Figure \ref{fig:gpt4v_failure_Case}, to determine the next action plan for the goal ``mixing the meat with yogurt'', the model must track several state changes of the yogurt container and spoon, identify the white yogurt paste on the meat, and recognize that further stirring is necessary. Even advanced GPT-4V could make wrong decisions.

This raises the question: \textbf{``How far are current MLLMs from achieving human-level planning?''} To investigate this, we introduce a comprehensive and rigorous benchmark called \textbf{EgoPlan-Bench} to evaluate the planning abilities of MLLMs in real-world scenarios from an egocentric perspective, which emulates human perception of the world. Unlike existing benchmarks \cite{fan2019egovqa, jia2022egotaskqa, li2023seed} that mainly focus on the \textbf{comprehension} ability of models, EgoPlan-Bench takes a step forward by systematically studying the \textbf{planning} capability of MLLMs. As shown in Figure \ref{fig:benchmark_overview}, to answer questions in EgoPlan-Bench correctly, the model must be capable of understanding open-form task goals, tracking long-horizon task progress, perceiving real-world environments from an egocentric view, and utilizing inherent world knowledge to reason about the next action plan in an end-to-end manner.

To construct EgoPlan-Bench, we leverage large-scale egocentric videos~\cite{Damen2022RESCALING, grauman2022ego4d} which authentically reflect daily human activities from the first-person perspective. We design a semi-automatic data construction pipeline (see Figure~\ref{fig:construction_pipeline}), and all the evaluation data has undergone strict manual verification. 
Our benchmark exhibits three main characteristics: 
\textbf{1) Realism of Tasks:} The tasks are extrapolated from authentic real-world videos, offering a closer reflection of daily human needs and showcasing greater variety than artificially designed tasks.
\textbf{2) Diversity of Action Plans:} The benchmark involves a diverse set of action plans, requiring interaction with hundreds of different objects and extending beyond basic manipulation skills such as picking and placing items.
\textbf{3) Intricacy of Visual Observations:} The visual observations come across various real-world scenes, where objects vary in appearance, state, and placement. Besides, the visual inputs can span extensive periods, making it challenging for monitoring task progress and detecting critical changes in object states.

We evaluate a total of 28 MLLMs on our benchmark. The results indicate that our benchmark poses significant challenges for existing MLLMs, and there is still a long way to go before these models evolve into human-level task planners. We further construct an instruction-tuning dataset \textbf{EgoPlan-IT}, specialized for enhancing human-level planning. The model fine-tuned on this dataset demonstrates robust performance gains on the proposed benchmark, verifying the effectiveness of our data.

In summary, our main contributions include: 1) We introduce an MLLM benchmark EgoPlan-Bench for human-level planning in real-world scenarios, which features realistic tasks, diverse actions, and intricate visual observations.
2) We evaluate a wide range of MLLMs and find that our benchmark poses significant challenges for them. We further provide a detailed analysis of the possible reasons. 
3) We construct an instruction-tuning dataset EgoPlan-IT specifically for human-level planning, which effectively improves model performance on EgoPlan-Bench. 
4) We have released all the codes and data, and are consistently maintaining a benchmark leaderboard to advance future research.

\section{Related Work}

\noindent \textbf{Multimodal Large Language Models (MLLMs).} Recent advancements in Large Language Models (LLMs)~\cite{ouyang2022training_instructGPT, touvron2023llama, chiang2023vicuna, chung2022scaling_flant5} have sparked interest in developing MLLMs that combine LLMs' language capabilities with multi-modal perception abilities~\cite{2023GPT4VisionSC,team2023gemini}. Typical research integrates representations from pretrained visual encoders into LLMs through pretraining on image-text datasets~ \cite{li2023blip2, dai2023instructblip, zhu2023minigpt4, liu2023visual_llava, liu2023llava1.5, ye2023mplugowl2, lu2024deepseek, young2024yi}. Recent studies extend this approach to incorporate video inputs~\cite{zhang2023video,li2023videochat,maaz2023videochatgpt}. Some works even advance MLLMs to include visual generation abilities~\cite{ge2023planting, ge2023making,ge2024seed}.

\noindent \textbf{MLLM Benchmarks.}
Numerous benchmarks have been proposed to assess MLLM performance recently. However, general-purpose benchmarks \cite{fu2023mme,yin2024lamm,li2023seed,liu2023mmbench,yu2023mm,yue2023mmmu,xu2023lvlm,ying2024mmt} allocate only a small proportion to egocentric vision, focusing mainly on comprehension capabilities. Specialized benchmarks such as EgoThink~\cite{cheng2023can} and PCA-Eval~\cite{chen2023towards} address egocentric thinking and embodied decision-making, respectively, but are constrained by limited number of handcrafted questions for planning and reliance on single-image visual observation.

\noindent \textbf{Anticipation and Planning in Videos.}
Human-level Planning (HP) is closely related to future anticipation and planning in videos. Future anticipation involves predicting future conditions using a video clip as context~\cite{sener2019zero,Damen2022RESCALING,grauman2022ego4d,abu2021long,furnari2017next,liu2022joint,lotter2016deep,villegas2017decomposing,mendonca2023structured}. While action anticipation shares similarities with planning, it does not account for specific goals. In contrast, goal-oriented action prediction is essential for real-world applications. Procedure planning~\cite{chang2020procedure} predicts actions using start and end images, but using images for goal specification is less natural than language instructions. VPA~\cite{patel2023pretrained} takes a language-specified goal and progress video to predict task-advancing actions. 
However, VPA primarily uses instructional YouTube videos~\cite{zhukov2019cross,tang2019coin}, which often capture third-person views and have pre-designed scripts and edits, limiting their alignment with real-world, egocentric human experiences.
Furthermore, VPA has not been extensively benchmarked with MLLMs. Egocentric Action Frame Generation~\cite{lai2023lego} focuses on egocentric vision but primarily addresses short-term visual synthesis for atomic actions, whereas HP emphasizes long-horizon planning.

\noindent \textbf{Egocentric Videos.}
Egocentric videos provide a distinctive perspective for active engagement with the real world. Numerous egocentric video datasets have been developed~\cite{pirsiavash2012detecting, li2015delving, sigurdsson2018charades}, with large-scale daily lifelog datasets such as Epic-Kitchens~\cite{Damen2022RESCALING} and Ego4D~\cite{grauman2022ego4d} offering fine-grained information on human-object interactions. Existing egocentric video QA benchmarks primarily assess models' understanding of activities and objects in a given video demonstration\cite{fan2019egovqa,jia2022egotaskqa}. However, none address the specific needs of human-level planning, where a practical question and answer would involve a task goal instruction and a prediction of the next appropriate action plan. Song \textit{et al.}~\cite{song2024ego4d} recently introduced an extended set of goal-oriented activity labels for Ego4D but focused on understanding abilities rather than addressing the gap in human-level planning.

\section{EgoPlan-Bench}\label{sec:benchmark}
\begin{figure*}[!t]
    \centering
    \includegraphics[width=1\textwidth]{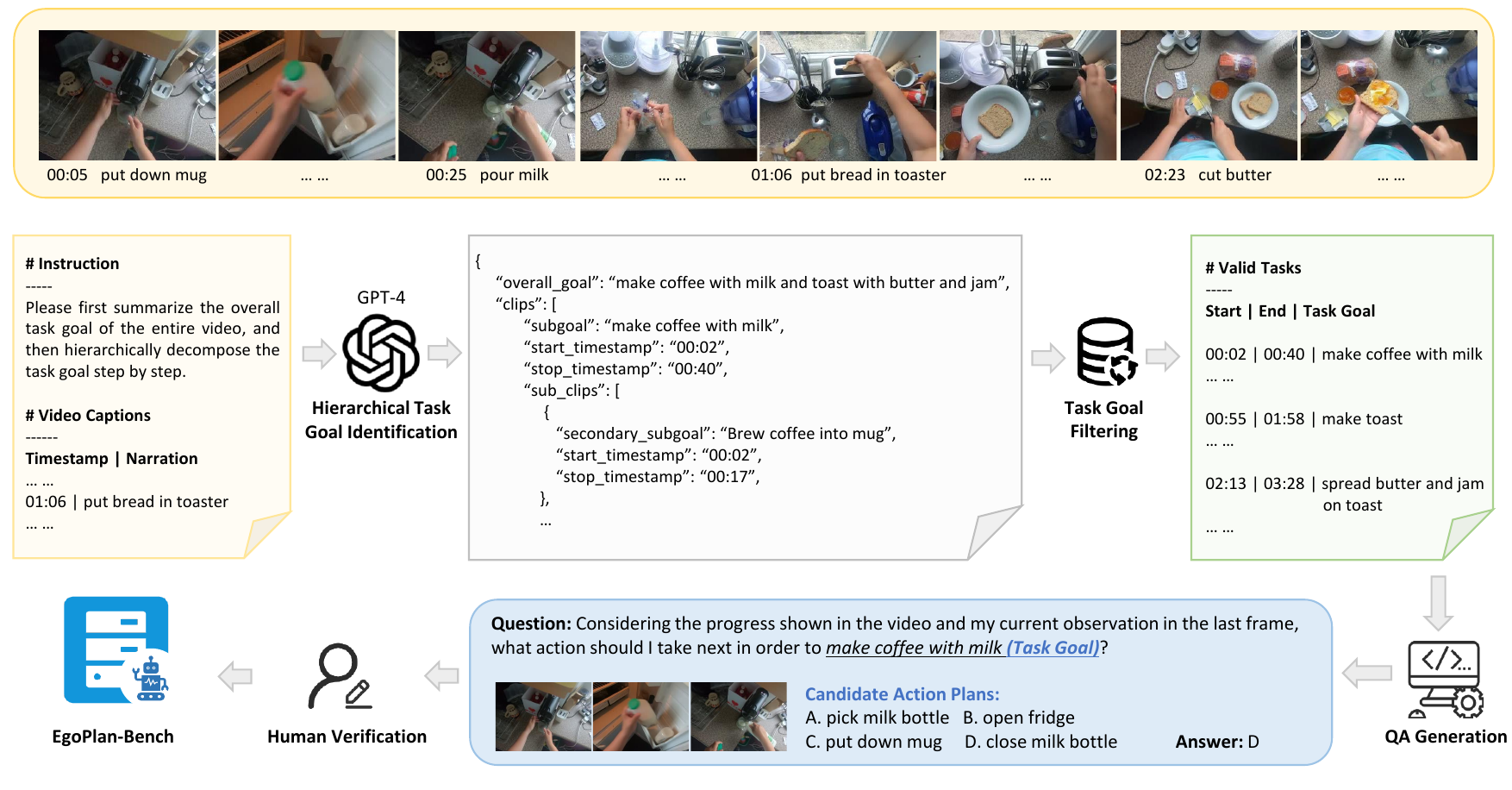}
    \caption{Overview of the construction pipeline for EgoPlan-Bench based on existing untrimmed egocentric videos with detailed action narrations. 
    (1) {We first leverage GPT-4 to identify task goals through hierarchical reasoning. } 
    (2) {We then filter task goals based on the requisite number of actions.}
    (3) The questions are designed in the form of multiple-choice, where the questions are automatically generated based on task goals, and the options are derived from different actions under the same task goal. (4) We employ human annotators to verify each question to ensure the benchmark quality.}
    \label{fig:construction_pipeline}
\end{figure*}
As illustrated in Figure \ref{fig:benchmark_overview}, given a language instruction specifying the task goal, Human-level Planning aims to decide the next appropriate action plan based on the visual input. The visual input includes a video showing a sequence of historical frames capturing the task progress, with the last frame indicating the current egocentric observation. 
The proposed EgoPlan-Bench contains 5K high-quality human-verified multiple-choice questions derived from  egocentric videos \cite{Damen2022RESCALING, grauman2022ego4d}. 
In Sec. \ref{sec:ego_videos}, we first introduce the video sources we used to build the benchmark. We illustrate how to automatically {extract} task goals from these videos in Sec. \ref{sec:goal_indentification} and structure questions based on the extracted task goals in Sec. \ref{sec:qa_generation}.
We then report the detailed data statistics in Sec. \ref{sec:data_statistics} and describe our evaluation strategy in Sec. \ref{sec:evaluation_strategy}.
The entire benchmark construction pipeline is demonstrated in Figure \ref{fig:construction_pipeline}.

\subsection{Egocentric Video Sources}\label{sec:ego_videos}
{
Unscripted, egocentric videos encompass a wide range of daily human tasks and diverse actions, making them an ideal and challenging medium for evaluating human-level planning.
We select two large-scale and well-annotated datasets, Epic-Kitchens \cite{Damen2022RESCALING} and Ego4D \cite{grauman2022ego4d}, as our video sources. 
In Epic-Kitchens, each video is annotated with dense narrations and start/end timestamps of actions. 
The narrations are expressed as brief verb-object phrases. 
Ego4D offers similar annotations, except that it only provides a single timestamp for each action and the narrations are expressed as brief sentences like \textit{``C does something''}.  We convert Ego4D narrations to verb-object phrases as Epic-Kitchens with GPT-4 \cite{openai2023gpt4} and derive the start and end timestamps for each action segment following EgoVLP~\cite{lin2022egocentric}.
}

\subsection{{Task Goal Extraction}}\label{sec:goal_indentification}

\noindent \textbf{Hierarchical Task Goal Identification.}
Egocentric videos, which capture tasks ranging from simple (involving fewer than three actions) to complex (entailing a long sequence of actions), present challenges in identifying task goals due to their varying lengths. To address this, we introduce a hierarchical reasoning strategy to identify task goals. 
As illustrated in Figure \ref{fig:construction_pipeline}, given the dense narrations of actions, we utilize GPT-4 to decompose video content into three levels: overall goal $\rightarrow$ subgoal $\rightarrow$ secondary subgoal. For instance, the goal to ``\textit{make coffee with milk and toast with butter and jam}'' is divided into the subgoals of ``\textit{making coffee with milk'' and ``making toast with butter and jam},'' with the former further broken down into secondary subgoals like ``\textit{brew coffee into mug}'' and ``\textit{merge coffee and milk}.'' This hierarchical approach allows GPT-4 to efficiently summarize tasks of varying complexity by organizing them into a structured framework, similar to chain-of-thought reasoning. Our findings suggest that prompting GPT-4 to generate all task goals at once may lead to difficulties due to the extensive information presented in longer videos. The detailed prompt can be found in Appendix~\ref{sec:detailed_prompt}.

\noindent \textbf{Task Goal Filtering.}
Given the variance in video lengths, the task hierarchy across different videos might not align consistently. Specifically, the complexity of the overall goal in a shorter video could be equivalent only to a secondary subgoal in a more extended video. 
To address this variance, we aggregate the overall goals, subgoals, and secondary subgoals extracted from various videos under the unified term ``task goals''. 
We then refine this set by filtering based on the requisite number of actions, retaining only those task goals that involve between 4 to 20 actions to maintain a moderate level of task complexity.

\subsection{\chenyi{Multiple-Choice Question Answering}}\label{sec:qa_generation}
Following the mainstream MLLM benchmarks \cite{li2023seed, liu2023mmbench, yue2023mmmu, ying2024mmt}, we evaluate human-level planning with multiple-choice questions, which facilitates the convenient computation of accuracy as an objective metric. Furthermore, such a question structure allows us to carefully design negative choices to control the quality and difficulty of our benchmark.

\noindent \textbf{Automatic QA Generation.}
With the filtered task goals, we use templates to automatically generate questions. 
As shown in the blue box in Figure \ref{fig:construction_pipeline}, the underlined words indicate the part to fill in with different task goals. 
With the same task goal, we can derive several question-answering (QA) pairs that differ in task progress and current observations. 
For example, given a task goal involving $N$ {actions}, we can obtain $N$ corresponding QA pairs with each action as the ground-truth answer.

\noindent \textbf{Visual Input Alignment.} 
We align the visual input with the ground-truth answer for each QA pair. 
The visual input contains the historical observations since the whole task begins and right before the action described by the ground-truth answer occurs. 
The last frame of the visual input indicates the current observation for planning the next action. 
To prevent the models from cheating with the clues from hand-object interaction (e.g., a hand touching a tap implies turning it on or off), we set the end timestamp of the visual input as 0.5 seconds preceding the actual next-action segment.

\noindent \textbf{Negative Choice Selection.} For each QA pair, we randomly sample three distinct actions with the same task goal, excluding the ground-truth action, as negative choices. This challenging setting demands a deep understanding of the environment state from dynamic visual input and world knowledge, such as action order dependency, rather than just the semantic meanings of task goals and actions, to discern the correct action plan. All candidate options appear reasonable for the given task goal if the visual modality is ignored. To maintain consistent granularity among options and ensure a unique correct answer, each multiple-choice question has been verified by human annotators.

\noindent \textbf{Human Verification.}
We employ 10 expert annotators with at least a bachelor's degree to meticulously verify the accuracy of each auto-generated multiple-choice question. Initially, annotators must pass a quality assurance phase by pre-annotating 50 samples according to comprehensive guidelines, which are then reviewed by two senior annotators.
During the main annotation phase, annotator pairs independently select the most suitable answers for each question based on the task goal and visual cues. They also flag questions with unclear visuals or ambiguous options for removal. We only retain questions with unanimous annotator agreement and alignment with the ground-truth answer, resulting in 5K high-quality evaluation instances closely correlated with visual observations. The detailed guidelines for human annotators are provided in Appendix~\ref{sec:annotation_guidlines}.

\subsection{\chenyi{Data Statistics}}\label{sec:data_statistics}
\begin{figure}[!t]
  \centering
  \begin{minipage}{0.5\linewidth}
    \centering
    \includegraphics[scale=0.42]{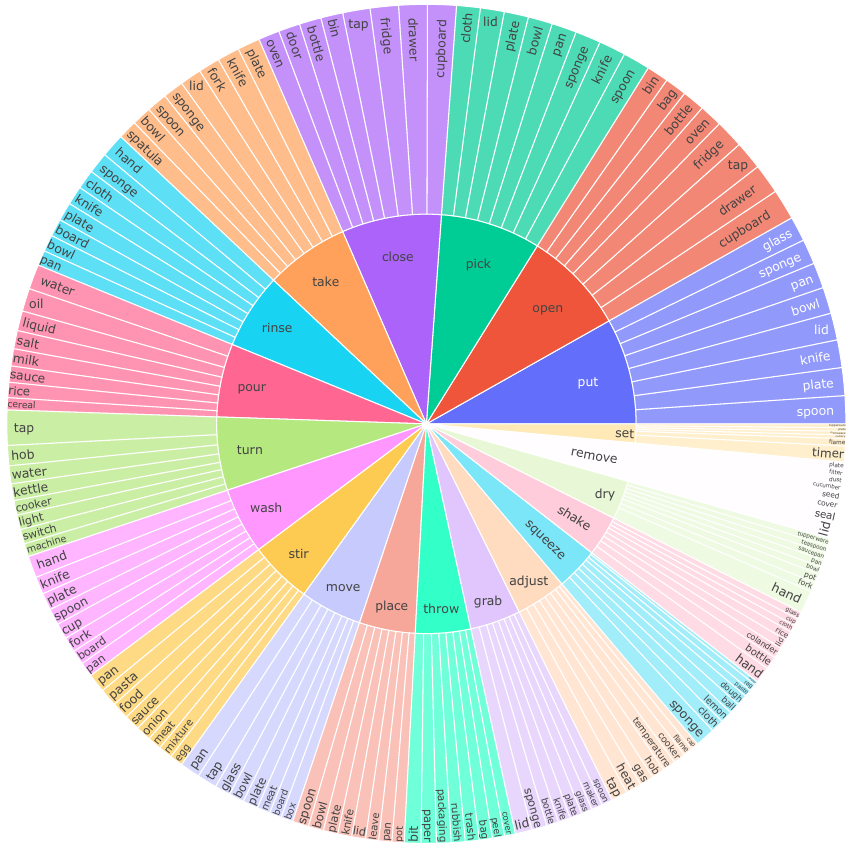}
    \caption{{Top 20 verbs with top 8 related objects in EgoPlan-Bench candidate action plans.}}
    \label{fig:action_statistics}
  \end{minipage}
  \hfill
  \begin{minipage}{0.45\linewidth}
    \centering
    \vspace{6.2mm}
    \includegraphics[scale=0.17]{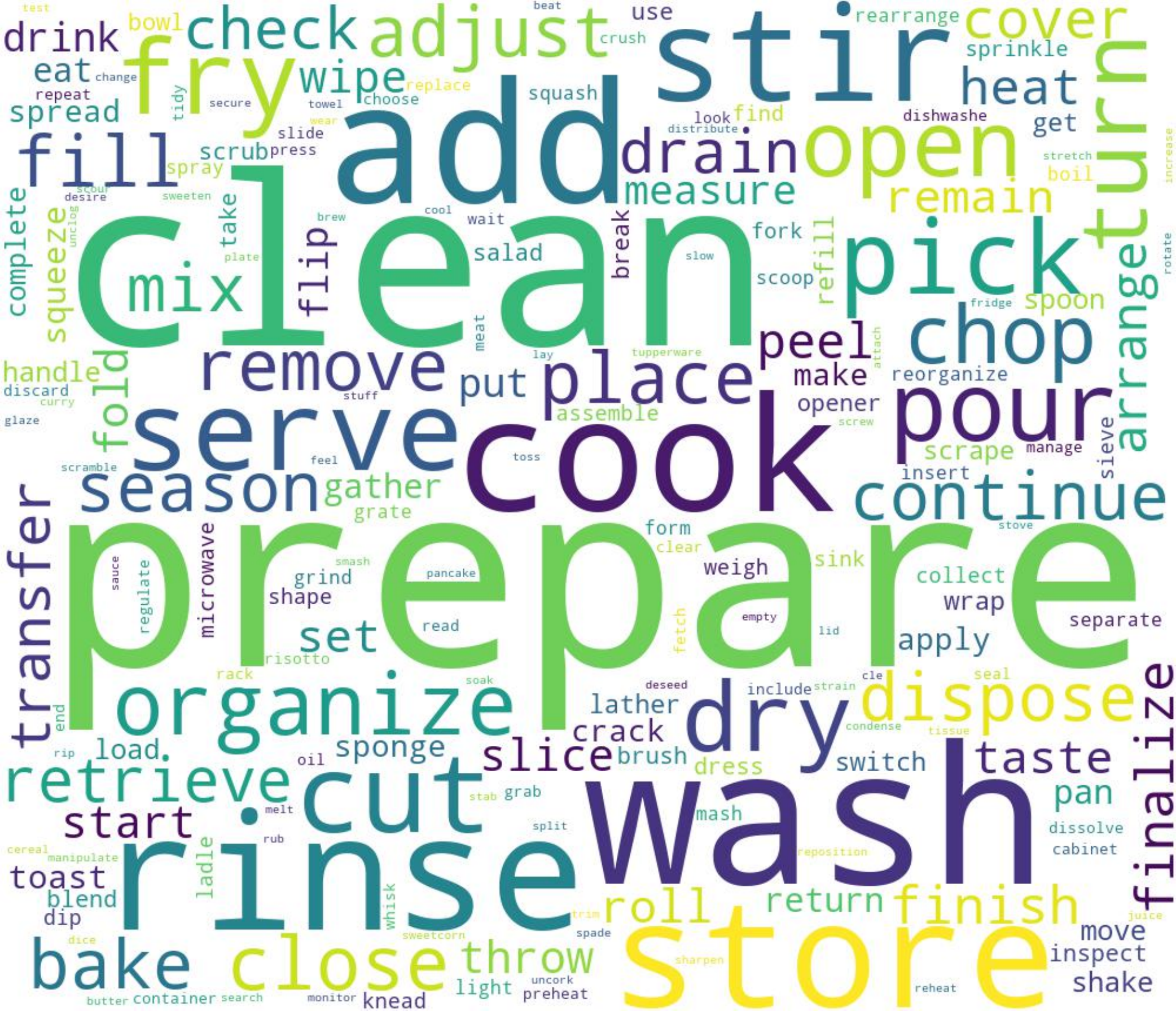}
    \vspace{2.3mm}
    \caption{{Wordcloud of task goals involved in EgoPlan-Bench questions.}}
    \label{fig:task_verb_wordcloud}
  \end{minipage}
  \label{fig:short}
\end{figure}

The evaluation data of EgoPlan-Bench, detailed in Table \ref{tab:data_statistics}, includes 4,939 multiple-choice questions covering 3,296 task goals. Figure \ref{fig:task_verb_wordcloud} shows the distribution of task goals.
Overall, there are 3,185 action plans (verb-object phrases) from the questions' candidate options, involving 234 distinct root verbs and 558 unique objects.
The most common action plans are illustrated in Figure \ref{fig:action_statistics}, demonstrating the benchmark's variety of action plans. 
Additionally, the visual observations of these questions originate from egocentric videos taken across 419 different scenes. 
This diversity in task goals, candidate actions, and visual scenes underscores the challenges presented by our benchmark.
We divide the evaluation data into two subsets: EgoPlan-Val for validation and EgoPlan-Test for testing. To ensure a fairer public comparison of model performance and to enable future opportunities for hosting related challenges, the ground-truth answers of EgoPlan-Test are kept private.

\subsection{Evaluation Strategy}\label{sec:evaluation_strategy}

Inspired by \cite{brohan2023can, dai2023instructblip, brown2020gpt3, lin2021truthfulqa, li2023seed}, we use the closed-set answer ranking strategy based on language completion distribution, rather than directly matching a model’s free-form generation results with candidate options. 
From the output distribution of MLLMs, we can derive the likelihood $p(a|v,l)$ that an MLLM generates the content of each candidate action plan $a$, given the visual input $v$ and the user instruction $l$ as the multimodal context.

This approach avoids the influence of option sequence in the prompt on the model's performance. 
More importantly, it is closer to the practical setting. 
An MLLM that achieves better performance on our benchmark indicates that it can output precise next-action plans with higher probability.

\section{\chenyi{Evaluation Results}}
\subsection{Models}
We select 28 prominent MLLMs for a comprehensive evaluation, including BLIP-2~\cite{li2023blip2}, InstructBLIP~\cite{dai2023instructblip}, InstructBLIP Vicuna~\cite{dai2023instructblip}, LLaVA~\cite{liu2023visual_llava}, MiniGPT-4~\cite{zhu2023minigpt4}, VPGTrans~\cite{2023vpgtrans},  MultiModal-GPT~\cite{gong2023multimodalgpt}, Otter~\cite{li2023otter}, OpenFlamingo~\cite{openflamingo}, LLaMA-Adapter V2~\cite{gao2023llamaadapterv2}, GVT~\cite{wang2023gvt}, mPLUG-Owl~\cite{ye2023mplugowl}, Kosmos-2~\cite{peng2023kosmos}, Qwen-VL-Chat~\cite{bai2023qwen}, LLaVA1.5~\cite{liu2023llava1.5}, VideoChat~\cite{li2023videochat}, Video-ChatGPT~\cite{maaz2023videochatgpt}, Valley~\cite{luo2023valley}, Video-LLaMA~\cite{zhang2023video}, SEED-LLaMA~\cite{ge2023making}, CogVLM~\cite{wang2023cogvlm}, DeepSeek-VL-Chat~\cite{lu2024deepseek}, mPLUG-Owl-2~\cite{ye2023mplugowl2}, Yi-VL~\cite{young2024yi}, SEED-X~\cite{ge2024seed},  Gemini-Pro-Vision~\cite{team2023gemini},  XComposer~\cite{zhang2023internlm}, {GPT-4V~\cite{2023GPT4VisionSC}}.

\begin{wraptable}{r}{0.5\textwidth}
    \centering
    \small
    \vspace{-12pt}
    \caption{Performance of MLLMs on EgoPlan-Val.}\label{tab:evaluation_results}
    \begin{tabular}{c c c}
         \toprule
         \textbf{Model} & \textbf{LLM} &   \textbf{Acc\%} \\
         \midrule
         BLIP-2 &Flan-T5-XL & 26.71 \\
         InstructBLIP &Flan-T5-XL  & 28.09  \\
         InstructBLIP Vicuna &Vicuna-7B  & 26.53    \\
         LLaVA &LLaMA-7B & 27.00  \\
         MiniGPT-4 &Vicuna-7B & 28.11   \\
         VPGTrans &LLaMA-7B & 27.38   \\
         MultiModal-GPT &Vicuna-7B& 27.81  \\
         Otter &LLaMA-7B  & 28.08    \\
         OpenFlamingo &LLaMA-7B  & 27.67  \\
         LLaMA-Adapter V2 &LLaMA-7B   & 27.81    \\
         GVT &Vicuna-7B  &  27.87    \\
         mPLUG-Owl &LLaMA-7B & 27.63   \\ 
         Kosmos-2 & Decoder only 1.3B  & 26.97 \\
         Qwen-VL-Chat & Qwen-7B & 27.69      \\
         LLaVA-1.5 &Vicuna-7B   &  27.81    \\
         VideoChat &Vicuna-7B  & 27.51   \\
         Video-ChatGPT &LLaMA-7B   &  27.33   \\
         Valley &LLaMA-13B  &  27.27  \\
         Video-LLaMA & LLaMA2-Chat-7B  &  28.58 \\
         SEED-LLaMA & LLaMA2-Chat-13B    & 29.93     \\
         CogVLM & Vicuna-7B & 27.48 \\
         DeepSeek-VL-Chat & DeepSeek-LLM-7B & 27.57 \\
         mPLUG-Owl-2 & LLaMA2-7B & 27.84 \\
         Yi-VL & Yi-6B & 28.67 \\
         Gemini-Pro-Vision & - & 30.46 \\
         SEED-X & LLaMA2-Chat-13B & 31.07 \\
         XComposer & InternLM-7B   &  {37.17}   \\
         GPT-4V & {-} & \textbf{37.98} \\
         \bottomrule
    \end{tabular}
    \vspace{-13pt}
\end{wraptable}

\subsection{\chenyi{Performance Analysis}}

\begin{figure}
\vspace{-10pt}
    \begin{minipage}{0.45\linewidth}
    \centering
    \includegraphics[scale=0.45]{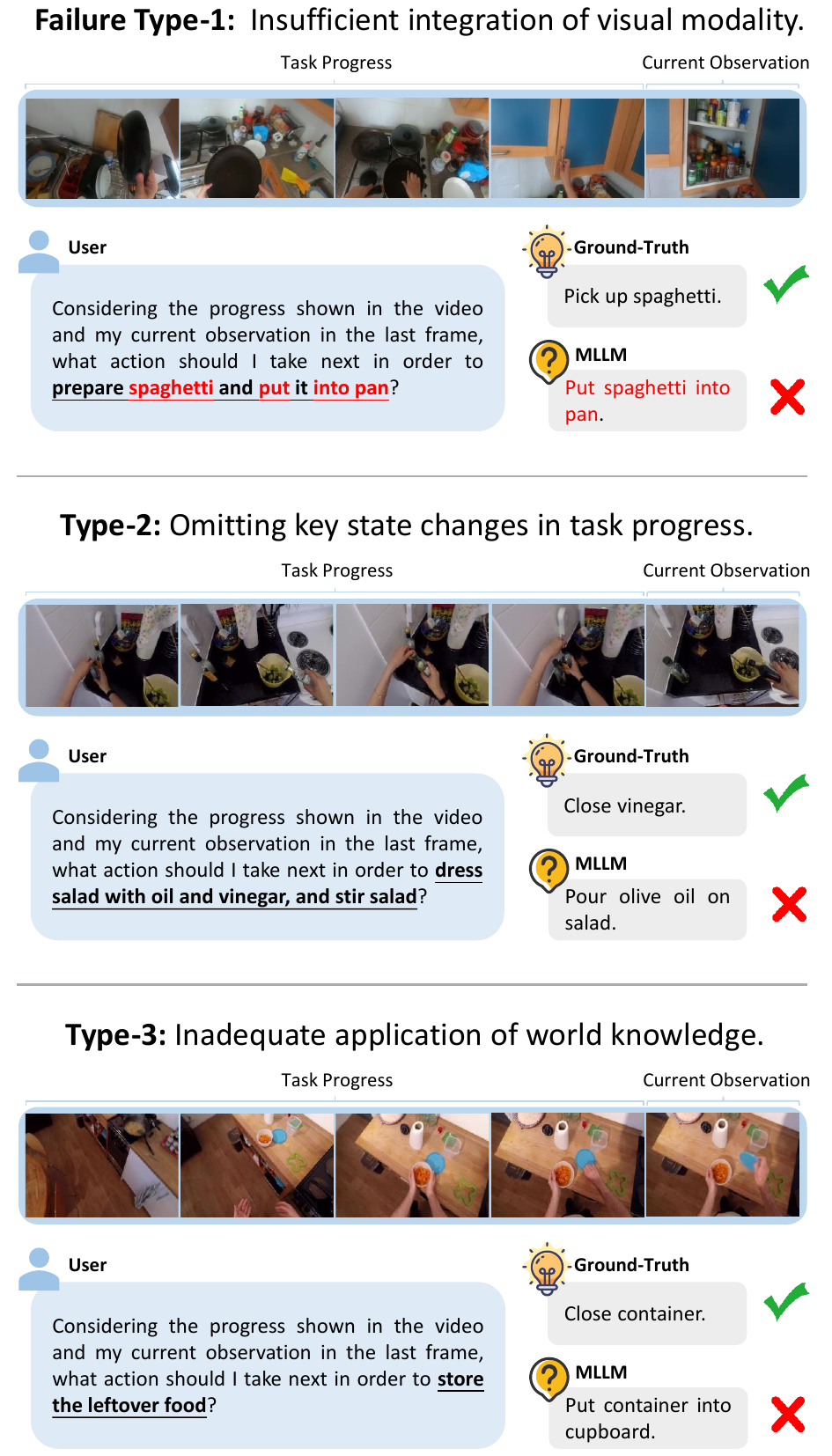}
    \caption{Common failure cases of the evaluated MLLMs on EgoPlan-Val.}
    \label{fig:detailed_case_study}
    \end{minipage}
    \hfill
    \begin{minipage}{0.45\linewidth}
    \centering
    \vspace{15pt}
    \includegraphics[scale=0.26]{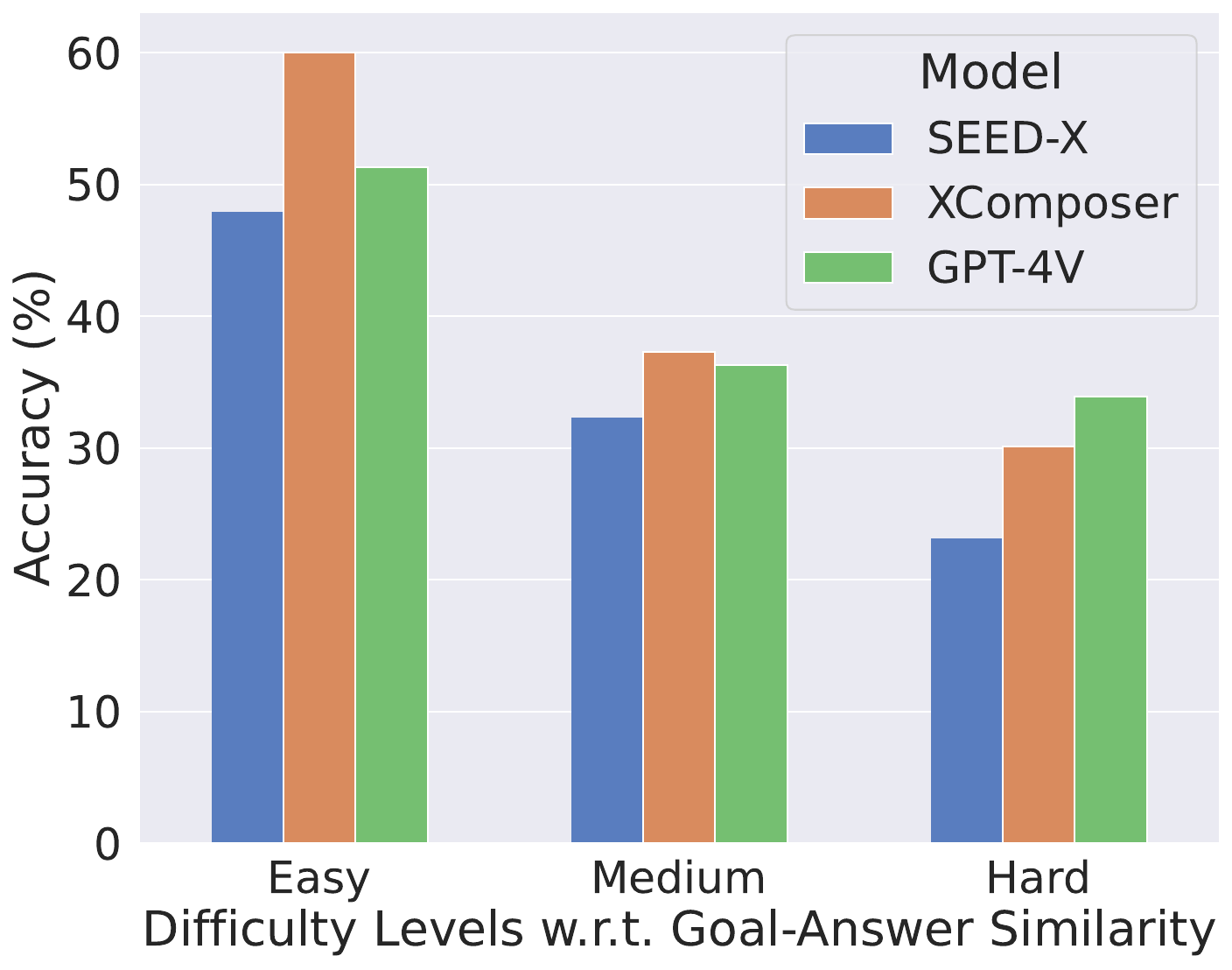}
    \caption{The influence of goal-answer similarity on the top-3 model performance.}
    \label{fig:analysis_goal_answer_similarity}
    \vspace{35pt}
    \includegraphics[scale=0.26]{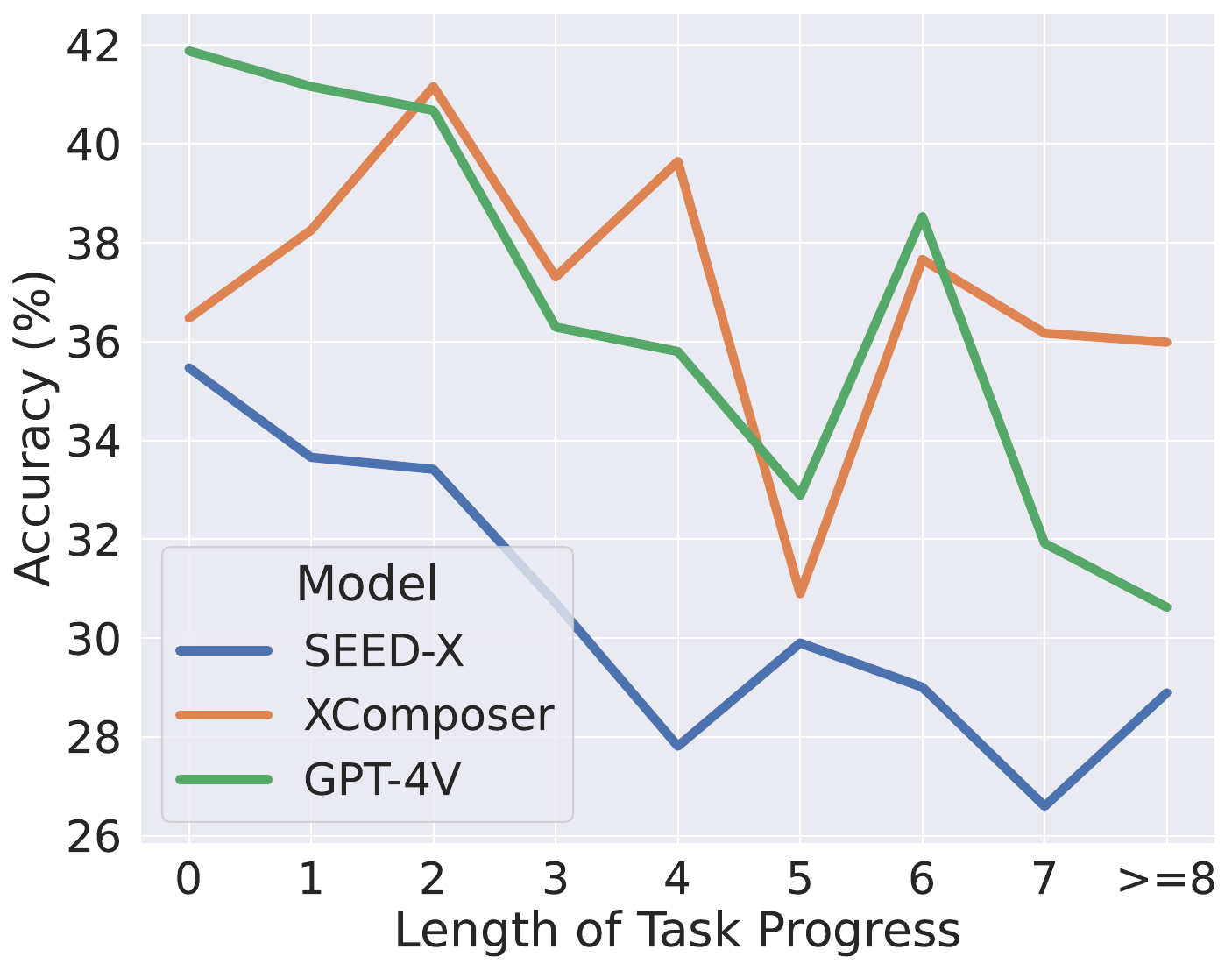}
    \caption{{The influence of task progress length on the top-3 model performance.}}
    \label{fig:analysis_progress_len}
    \end{minipage}
\end{figure}

In this paper, we evaluate various MLLMs on the EgoPlan-Val set, with results on the EgoPlan-Test set available in Appendix~\ref{sec:egoplan_test_results}. Table \ref{tab:evaluation_results} shows that most models perform only slightly better than random guessing, with the best model (GPT-4V) achieving just 37.98\% accuracy. 
This indicates that current MLLMs are not yet capable of mature human-level planning. 
In Appendix~\ref{sec:case_study_gpt4v}, we provide an in-depth case study with GPT-4V. 
Here, we summarize three main reasons for this limitation:

\textbf{(1) Insufficient integration of visual modality:} Most MLLMs, adapted from Large Language Models, excel in semantic reasoning but struggle with real-world observations necessary for human-level planning. Consequently, they often select options semantically closer to the task goal while ignoring visual cues (see Failure Type-1 in Figure~\ref{fig:detailed_case_study}). 
When we categorized EgoPlan-Val instances into three difficulty levels based on semantic similarity (calculated by BertScore~\cite{zhang2019bertscore}) between the task goal and the ground-truth answer, the top three models perform significantly better on easier questions with higher goal-answer similarity (Figure~\ref{fig:analysis_goal_answer_similarity}).

\textbf{(2) Omitting key state changes in task progress:} Long-horizon tasks require models to monitor state changes of key objects, necessitating both long-context memory and high-resolution visual observation processing. As task progress lengthens (measured by the number of historical actions), it becomes harder to capture state changes, and the model performance generally decreases (Figure~\ref{fig:analysis_progress_len}).  For example, in Failure Type-2 (Figure~\ref{fig:detailed_case_study}), the model plans to "pour olive oil on salad" without recognizing that this action has already been completed.

\textbf{(3) Inadequate application of world knowledge:} Human-level planning also relies on world knowledge. 
In Failure Type-3 (Figure~\ref{fig:detailed_case_study}), to "store leftover food," "closing the container" is essential for preserving freshness before "putting it somewhere". However, current MLLMs struggle to integrate their inherent world knowledge into the planning process as seamlessly as humans.

\begin{table}[!t]
    \centering
    \small{
    \setlength{\tabcolsep}{1mm}{
    \caption{Statistics of the evaluation data of EgoPlan-Bench.}
    \begin{tabular}{c c c c c c c}
    \toprule
   split & {\#questions} & {\#task goals} & {{\#action plans (verb+object)}} & \#verbs & {\#objects} & {\#scenes} \\
    \midrule
    EgoPlan-Val & 3,355 & 2,407 & 2,538 & 202 & 496 & 279 \\   
    EgoPlan-Test & 1,584 & 902 & 1,492 & 160 & 313 & 317 \\
    Total & 4,939 & 3,269 & 3,185 & 234 & 558 & 419 \\
    \bottomrule
    \end{tabular}\label{tab:data_statistics}}}    
\end{table}

\section{Enhancing Human-Level Planning through Instruction Tuning}

Given the suboptimal performance of the evaluated MLLMs on EgoPlan-Val, 
we investigate enhancing the human-level planning capabilities of MLLMs through instruction-tuning. Specifically, we construct an instruction-tuning dataset, EgoPlan-IT, to align MLLMs with real-world needs of planning. 
In this section, we first describe the data preparation process in Sec.  \ref{sec:data_preparation}. 
Then we introduce the training objectives in Sec. \ref{sec:training_objective}. 
In Sec. \ref{sec:experiments}, we demonstrate the effectiveness of our data and instruction-tuning strategy through experiments.

\subsection{\chenyi{Data Preparation}}\label{sec:data_preparation}
\noindent \textbf{Construction of EgoPlan-IT.} Following the pipeline in Sec. \ref{sec:benchmark}, we additionally tailor an instruction-tuning dataset EgoPlan-IT, which contains 50K QA pairs for human-level planning, as shown in the bottom half of Figure \ref{fig:benchmark_overview}. 
To assess the robustness of our instruction-tuning approach, we exclusively utilize videos  from Epic-Kitchens \cite{Damen2022RESCALING} for the creation of this dataset. 
This allows us to assess the enhanced model's performance on both the in-domain (Epic-Kitchens) subset, comprising 2,432 samples from EgoPlan-Val, and the out-of-domain (Ego4D) subset, consisting of 923 samples from EgoPlan-Val.
In order to increase the efficiency for constructing this dataset, we do not include human annotations in this process.

\noindent \textbf{Auxiliary Data for Action Recognition.}
To reduce the learning difficulty for the model, we also construct an auxiliary dataset for action recognition based on the visual inputs of the 50K training samples from EgoPlan-IT. For example in the bottom half of Figure \ref{fig:benchmark_overview}, the instruction of the auxiliary sample would be a simple query \textit{``Can you enumerate the actions in the video, describing each with a short verb-noun combination?''}, and the response is the concatenation of corresponding action narrations such as \textit{``Open yogurt, scoop yogurt, put yogurt on meat, put down yogurt.''}

\noindent \textbf{Increasing Instruction Diversity.}
In order to increase the diversity of instructions, we also include an additional collection of 164K instruction data, which integrates the 150K  image-instruction dataset from LLaVA \cite{liu2023visual_llava}, the 3K image-detail-description dataset from MiniGPT-4 \cite{zhu2023minigpt4} and the 11K video-instruction dataset from VideoChat \cite{li2023videochat}.

\subsection{Training Objectives} \label{sec:training_objective}

We use Video-LLaMA \cite{zhang2023video} as a baseline MLLM for investigation, as shown in Figure~\ref{fig:model_framework}. A detailed description of the model framework can be found in Appendix~\ref{sec:model_framework}. 
To fine-tune the model, we mix the 50K specialized EgoPlan-IT data together with the 50K auxiliary data for Action Recognition and the 164K diverse instruction data. The autoregressive loss function is defined as:
\begin{equation}
    L_{ft} = -\sum_t \log P(a_t | \boldsymbol{v}, \boldsymbol{l}, \boldsymbol{a}_{<t}; \boldsymbol{\Theta}),
\end{equation}
where $\boldsymbol{v}$ and $\boldsymbol{l}$ denote the given visual observations and the language instruction, and  $\boldsymbol{a}_{<t}$ signifies the sequence of answer tokens preceding the current prediction token  $a_t$. $\boldsymbol{\Theta}$ is the trainable model parameters with LoRA \cite{hu2021lora}. 
Inspired by RRHF \cite{yuan2023rrhf}, we add a contrastive loss to the fine-tuning loss:
\begin{equation}
    L = L_{ft} + L_{ct},
\end{equation}
\begin{equation}
    L_{ct} = max(0, s_{neg}-s_{pos}),
\end{equation}
\begin{equation}
    s = \frac{\sum_t \log P(a_{t}| \boldsymbol{v}, \boldsymbol{l}, \boldsymbol{a}_{<t}; \boldsymbol{\Theta})}{||\boldsymbol{a}||}.
\end{equation}
Here, for each instruction-following pair, a random action plan distinct from the ground-truth answer is selected under the same task goal to serve as a `negative' answer. The model is constrained to ensure that its score for the `positive' answer ($s_{pos}$) exceeds that for the `negative' answer ($s_{neg}$). 
This operation encourages a more balanced consideration of both the language and visual modalities. We describe the model implementation details in Appendix~\ref{sec:implementation_details}.

\begin{figure}[!t]
    \centering
    \includegraphics[scale=0.5]{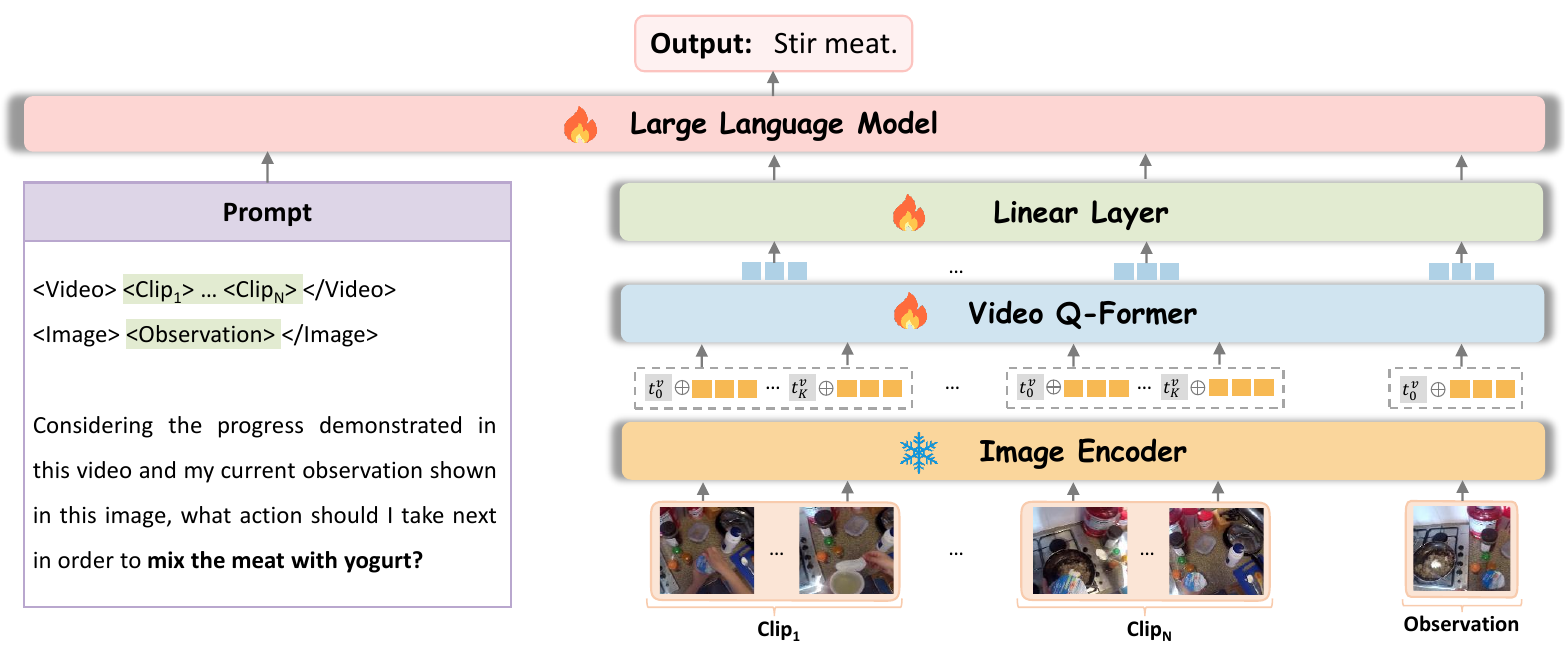}
    \caption{
    The diagram of enhancing the baseline model Video-LLaMA with EgoPlan-IT.}
    \label{fig:model_framework}
\end{figure}

\subsection{Experiments}\label{sec:experiments}
\noindent \textbf{Main Results.}
The results in Table \ref{tab:instruction_tuning_results} clearly indicate that the baseline model (Video-LLaMA) tuned on our data outperforms the previous state-of-the-art model, 
{GPT-4V}, 
by a considerable margin. Notably, it shows a remarkable 23.25\% increase in accuracy over its vanilla version across the overall validation set. Furthermore, the enhanced model demonstrates robust domain transfer capabilities, outperforming the vanilla version by 13.98\% on the out-of-domain validation subset. This indicates that the planning experience gained through instruction-tuning on our EgoPlan-IT data is effectively transferable to new environments.

\begin{wraptable}{r}{0.5\textwidth}
    \centering
    \small
    \vspace{-19pt}
    \caption{Performance of the baseline model (Video-LLaMA) tuned on EgoPlan-IT. ID and OD are short for in-domain and out-of-domain.}
    \begin{tabular}{l c c c}
    \toprule
    \textbf{Model} & \textbf{ID} & \textbf{OD} & \textbf{All} \\
    \midrule
    GPT-4V &  38.40 & 36.90 & 37.98 \\
    Video-LLaMA & 27.88 &  30.44 &  28.58 \\
    \midrule
    Enhanced Video-LLaMA & \textbf{54.65} & \textbf{44.42} & \textbf{51.83} \\
    w/o LoRA & 53.58 & 42.58  & 50.55 \\
    w/o Contrastive Loss & 52.14 & 40.52  & 48.94  \\
    w/o Diverse Instructions & 53.95 & 43.12  & 50.97 \\
    w/o Action Recognition & 52.34 &  {44.31} &  50.13 \\
    \bottomrule
    \end{tabular}
    \label{tab:instruction_tuning_results}
\end{wraptable}

\noindent \textbf{Ablations.} 
Our analysis of the ablated versions of the enhanced model reveals several key insights. Notably, employing the LoRA strategy, integrating contrastive loss, diversifying instructions, and fine-tuning with auxiliary action recognition data consistently enhance performance across domains. Contrastive loss contributes the most to the model's improvement. Furthermore, fine-tuning the model with in-domain action recognition auxiliary data boosts in-domain performance but does not significantly improve domain transferability.

\section{Conclusion}
In this work, we introduce EgoPlan-Bench, a sophisticated benchmark to evaluate Multimodal Large Language Models (MLLMs) in human-level planning, where a model is expected to plan step-by-step actions by considering task progress, current visual observations, and open-form task goals.
The evaluation results of various models reveal that
current MLLMs have not yet evolved into generalized human-level planners. 
We further construct EgoPlan-IT, an instruction-tuning dataset, to facilitate the learning of task planning experience from human videos. 
{The model tuned on EgoPlan-IT exhibits a significant performance improvement on our benchmark. We have released all the codes and datasets, and continuously maintain a leaderboard to promote future research.

\clearpage
\printbibliography{}
\newpage
\clearpage
\setcounter{page}{1}
\appendix

\section{Details for Data Curation}
\subsection{Hierarchical Task Goal Identification}\label{sec:detailed_prompt}

In Figure \ref{fig:prompt_for_task_decomposition}, we present the detailed prompt designed for the hierarchical identification of task goals with GPT-4. Initially, GPT-4 is prompted to meticulously analyze narrations from an ego-centric video, focusing on discerning the sequence and timing of actions. This analysis includes identifying temporal relationships, conditions preceding and succeeding actions, interactions with objects, and possible alterations within the environment or specific functional areas. 
Following this analysis, GPT-4 must distill the overarching goal of the video into an ``overall goal''. This overall goal is then broken down into a hierarchy of ``subgoals'' and, further, into more granular ``secondary subgoals'', facilitating a step-by-step reasoning process for identifying task goals.

\begin{figure}[!h]
    \centering
    \includegraphics{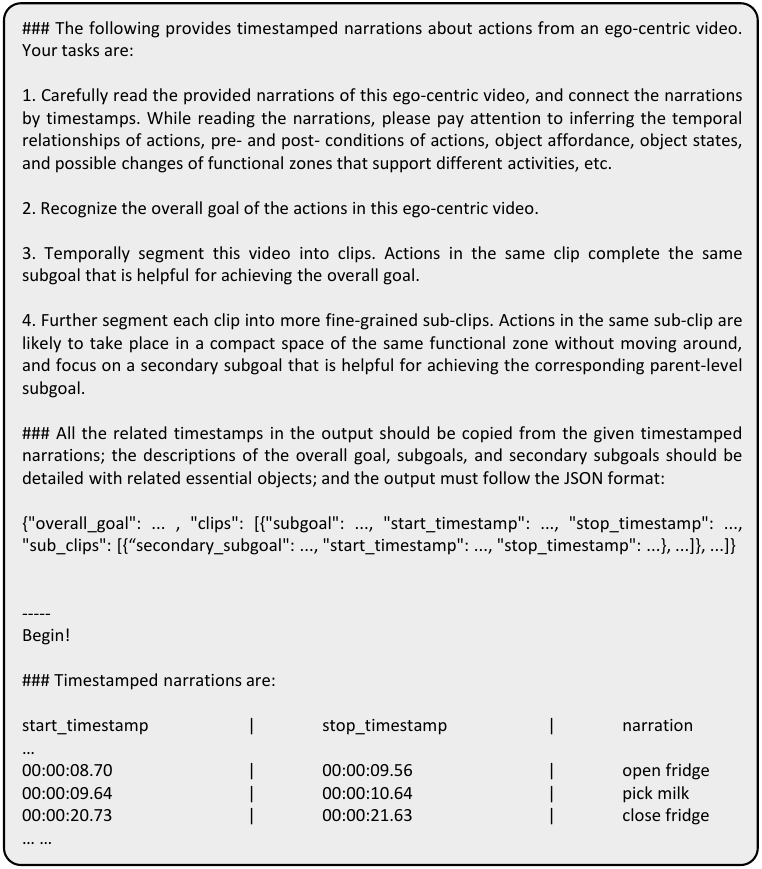}
    \caption{The detailed prompt for task goal identification with GPT-4.}
    \label{fig:prompt_for_task_decomposition}
\end{figure}

\subsection{Payment for Human Annotators}\label{sec:payment_for_annotators}
We have employed 10 annotators with at least a bachelor's degree to work on our datasets. They receive an hourly wage of approximately 7 USD, and the overall expense for their services amounts to about 2,500 USD.

\subsection{Instructions for Human Verification}\label{sec:annotation_guidlines}

\textbf{Annotation Task Overview}

Each annotation example consists of four components:

\begin{enumerate}
    \item A Question describing the Task Goal, which may involve baking, doing laundry, preparing ingredients, tidying up the room, etc.
    \item An Egocentric Video Clip representing the Task Progress, which captures a series of completed activities related to the current task.
    \item A Current Observation image that illustrates the environmental state from the egocentric view before performing the next action.
    \item Four Candidate Choices outlining potential action plans.
\end{enumerate}
Your task is to infer the task progress by considering the Task Goal from the Question and the Task Progress video. Then, select the most reasonable next action plan from the Candidate Choices that effectively advances the Task Goal and is feasible within the Current Observation's environment.

\textbf{Annotation Process}

\begin{enumerate}
    \item Begin by reading and understanding the Question, focusing on the task goal it addresses.

    \item Carefully watch the Task Progress video, identifying state changes of key objects related to the task goal, and infer the task progress.

    \item Examine the Current Observation image, identifying potential target objects for operation within the scene.

    \item Choose the most reasonable next action from the Candidate Choices (A$\sim$D) that advances the task and can be executed within the Current Observation. This action should:\\

    i) Align with the task goal specified in the Question, rather than completing unrelated tasks. \\
    ii) Involve target objects that are clearly visible within the Current Observation. \\
    iii) Prioritize actions that should occur earliest when multiple actions meet the conditions, based on the analysis of the sequence of actions.\\

    If none of the Candidate Choices (A$\sim$D) meet the above criteria, or if the task goal description is ambiguous, or if it is impossible to accurately assess the task progress, or if more than one reasonable option exists, select "E (Skip)".
    
\end{enumerate}

\section{Details for Evaluation Results}
\subsection{Model Performance on EgoPlan-Test}\label{sec:egoplan_test_results}

\begin{table}[!t]
    \centering
    \small
    \caption{Performance of MLLMs on EgoPlan-Test.}\label{tab:egoplan_test_results}
    \begin{tabular}{c c c}
         \toprule
         \textbf{Model} & \textbf{LLM} &   \textbf{Acc\%} \\
         \midrule
         BLIP-2~\cite{li2023blip2} &Flan-T5-XL & 27.90 \\
         InstructBLIP~\cite{dai2023instructblip} &Flan-T5-XL  & 25.19  \\
         InstructBLIP Vicuna~\cite{dai2023instructblip} &Vicuna-7B  & 26.64    \\
         LLaVA~\cite{liu2023visual_llava} &LLaMA-7B & 28.16  \\
         MiniGPT-4~\cite{zhu2023minigpt4} &Vicuna-7B & 30.93   \\
         VPGTrans~\cite{2023vpgtrans} &LLaMA-7B & 24.12   \\
         MultiModal-GPT~\cite{gong2023multimodalgpt} &Vicuna-7B& 30.43  \\
         Otter~\cite{li2023otter} &LLaMA-7B  & 30.87    \\
         OpenFlamingo~\cite{openflamingo} &LLaMA-7B  & 30.18  \\
         LLaMA-Adapter V2~\cite{gao2023llamaadapterv2} &LLaMA-7B   & 30.43    \\
         GVT~\cite{wang2023gvt} &Vicuna-7B  &  29.67    \\
         mPLUG-Owl~\cite{ye2023mplugowl} &LLaMA-7B & 31.31   \\ 
         
         Qwen-VL-Chat~\cite{bai2023qwen} & Qwen-7B & 31.06     \\
         LLaVA-1.5~\cite{liu2023llava1.5} &Vicuna-7B   &  29.80    \\
         VideoChat~\cite{li2023videochat} &Vicuna-7B  & 28.72   \\
         Video-ChatGPT~\cite{maaz2023videochatgpt} &LLaMA-7B   &  29.17   \\
         Valley~\cite{luo2023valley} &LLaMA-13B  &  30.11  \\
         Video-LLaMA~\cite{zhang2023video} & LLaMA2-Chat-7B  &  30.30 \\
         
         CogVLM~\cite{wang2023cogvlm} & Vicuna-7B & 31.06 \\
         DeepSeek-VL-Chat~\cite{lu2024deepseek} & DeepSeek-LLM-7B & 26.01 \\
         mPLUG-Owl-2~\cite{ye2023mplugowl} &LLaMA2-7B & 30.37   \\ 
         Yi-VL~\cite{young2024yi} & Yi-6B & 30.56 \\
         Gemini-Pro-Vision~\cite{zhang2023internlm} & -   &  {32.39}   \\
         SEED-X~\cite{ge2024seed} & LLaMA2-Chat-13B & 29.92 \\
         Xcomposer~\cite{zhang2023internlm} & InternLM-7B   &  {36.36}   \\
         GPT-4V~\cite{2023GPT4VisionSC} & {-} & \textbf{37.25} \\
         \bottomrule
    \end{tabular}
\end{table}

In Table \ref{tab:egoplan_test_results}, we present the performance of MLLMs on the held-out test set EgoPlan-Test. The results are consistent with that on the validation set EgoPlan-Val, showing that even the best models underperform on our benchmark. It further demonstrates that the proposed EgoPlan-Bench poses significant challenges for existing MLLMs in human-level planning.

\subsection{Case Study with GPT-4V}\label{sec:case_study_gpt4v}
Since GPT-4V achieves the best performance among all the evaluated MLLMs, we use it for a qualitative case study to delve deeper into why current MLLMs may underperform on our benchmark. 
As depicted in Figure \ref{fig:case_study}, we require GPT-4V to explicitly summarize the task progress and describe the current observation before planning the next action. 

Our observations indicate that while GPT-4V successfully identifies obvious actions, like \textit{``removing the lid of yogurt container''} or \textit{``measuring yogurt with a spoon,''} it might overlook more subtle state changes that are critical to the task's objective. For instance, GPT-4V fails to recognize that yogurt has been added to the meat, as indicated by the white paste visible on the meat post-measurement. Furthermore, even when the yogurt has been placed on the countertop in the final frame, GPT-4V erroneously maintains that the current observation shows \textit{``a hand holding a spoon containing yogurt.''} This misinterpretation of the visual input leads to incorrect plans, such as concluding that the next action is to \textit{``Add yogurt to meat,''} when this step has already been completed. 
This case study highlights a key area for improvement in MLLMs' processing of visual cues in the context of real-time task execution.

\begin{figure*}[!t]
    \centering
    \includegraphics[width=1\textwidth]{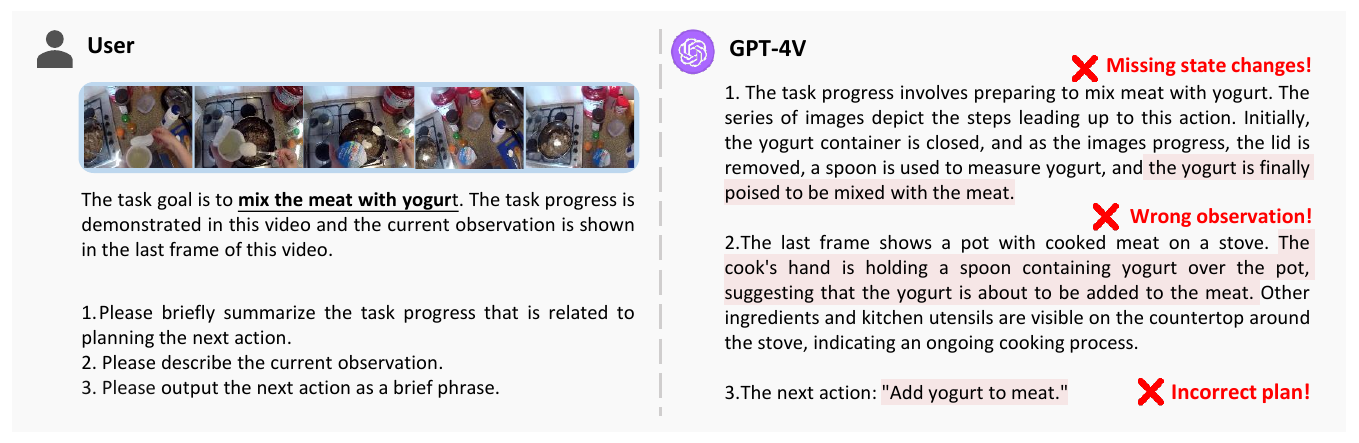}
    \caption{
    A case study with GPT-4V, where the model fails to detect the state changes of key objects in task progress, and struggles to recognize the target object in current observation, leading to the wrong prediction of action plan.
    }
    \label{fig:case_study}
\end{figure*}

Future directions may include: 1) developing advanced algorithms or incorporating sophisticated training datasets that highlight subtle visual changes, 2) enhancing contextual understanding of MLLMs to make more informed plans based on comprehensive visual input analysis relevant to specific task goals, and 3) integrating real-time feedbacks for adaptability in dynamic environments. These improvements would enhance MLLM performance in human-level planning and expand their applicability in real-world scenarios.

\section{Details for Instruction-Tuning}\label{sec:instruction_tuning_details}

\subsection{Model Framework}\label{sec:model_framework}

We use Video-LLaMA \cite{zhang2023video} as a baseline MLLM for investigation. 
As illustrated in Figure \ref{fig:model_framework}, the vision processor of Video-LLaMA comprises four components, including an Image Encoder to extract features from video frames, a position embedding layer for adding temporal information, a Video Q-former for aggregating frame representations, and a linear layer for producing video embedding vectors that match the dimensions of LLM token embeddings.

To enhance the model's ability to process visual inputs for human-level planning, we organize historical observations based on actions and retain the most recent $N (=4)$ clips. For each action segment, we choose $K (=8)$ uniformly spaced keyframes, creating an aggregated video representation. 
To emphasize the current observation, we treat it as a separate single-frame clip, extracting a unique representation for it. The token embeddings for placeholders \texttt{<$\text{Clip}_\text{i}$>} and \texttt{<Observation>} in the textual prompt are substituted with the embeddings corresponding to the $i$-th action segment and the current observation, respectively, within the LLM.


\subsection{Implementation Details}\label{sec:implementation_details}
The baseline model is built upon the Video-LLaMA (LLaMA2-Chat-7B) checkpoint which has undergone the second training stage\footnote{\href{https://huggingface.co/DAMO-NLP-SG/Video-LLaMA-2-7B-Finetuned}{https://huggingface.co/DAMO-NLP-SG/Video-LLaMA-2-7B-Finetuned}}. 
We further fine-tune the model on our data for 2 epochs using a batch size of 16 on 8 V100 GPUs, which takes 0.5 days. 
During this process, the Image Encoder is frozen and the language model is tuned using LoRA strategy, while the remaining components of the model are fully trainable.

\section{Limitation and Potential Negative Social Impacts}\label{sec:limitations_and_potential_negative_impacts}

While EgoPlan-Bench focuses on the planning capabilities of MLLMs in the context of visual observations and language instructions, it may not fully capture the complexities of real-world scenarios where other sensory inputs, such as auditory and tactile information, are also essential. The reliance on visual and language inputs could limit the applicability of the benchmark to broader real-world situations. 

As MLLMs become more capable of human-level planning, there is a risk of these models being used for malicious purposes, such as automating harmful actions or enabling surveillance and control. It is crucial to establish guidelines and regulations to prevent the misuse of these technologies.  Additionally, the development of AGI could lead to job displacement or exacerbate existing social inequalities if the benefits of these technologies are not distributed equitably.

\section{License}\label{sec:license}
EgoPlan-Bench is built upon two existing egocentric video sources: Epic-Kitchens~\cite{Damen2022RESCALING} and Ego4D~\cite{song2024ego4d}. The personal information in the used videos have been meticulously de-identified by their original dataset creators. Epic-Kitchens is available under the CC BY-NC 4.0 License, which allows us to utilize their data for non-commercial purposes. Ego4D is released under the Ego4D License, for which we have signed and received approval to download their data.

Our EgoPlan-Bench datasets, including EgoPlan-Val, EgoPlan-Test, and EgoPlan-IT, are also released under the CC BY-NC 4.0 License. We only release our annotated questions and answers for human-level planning, along with Video IDs for referencing corresponding video sources. Users should download the videos from their original websites using the Video IDs provided by us.

We employ Video-LLaMA~\cite{zhang2023video} as a baseline MLLM for instruction tuning. Video-LLaMA is licensed under the BSD 3-Clause License. We also open-source our code and the fine-tuned model under the BSD 3-Clause License, including a copy of Video-LLaMA's original license in our GitHub repository.

\end{document}